\documentclass[letterpaper]{article} % DO NOT CHANGE THIS
\usepackage{aaai2026}  % DO NOT CHANGE THIS
\usepackage{times}  % DO NOT CHANGE THIS
\usepackage{helvet}  % DO NOT CHANGE THIS
\usepackage{courier}  % DO NOT CHANGE THIS
\usepackage[hyphens]{url}  % DO NOT CHANGE THIS
\usepackage{graphicx} % DO NOT CHANGE THIS
\usepackage{natbib}  % DO NOT CHANGE THIS AND DO NOT ADD ANY OPTIONS TO IT
\usepackage{caption} % DO NOT CHANGE THIS AND DO NOT ADD ANY OPTIONS TO IT
\usepackage{algorithm}
\usepackage{algorithmic}
\usepackage{booktabs}
\usepackage{amsmath}
\usepackage{multirow}
\usepackage{makecell}
\usepackage{graphicx}
\usepackage{amsmath}
\usepackage{amsthm}
\usepackage{amssymb}
\usepackage{xcolor}
\usepackage{colortbl}
\usepackage{amsfonts}

\usepackage{newfloat}
\usepackage{listings}
\DeclareCaptionStyle{ruled}{labelfont=normalfont,labelsep=colon,strut=off} % DO NOT CHANGE THIS
\floatstyle{ruled}
\newfloat{listing}{tb}{lst}{}
\floatname{listing}{Listing}
\title{Reasoning Shapes Alignment: Investigating Cultural Alignment in \\ Large Reasoning Models with Cultural Norms}
\author{
    Yuhang Wang, 
    Yanxu Zhu, 
    Jitao Sang\thanks{Corresponding author.}
}
\affiliations{
    Beijing Key Laboratory of Traffic Data Mining and
Embodied Intelligence\\
    Beijing Jiaotong University\\
    yhangwang@bjtu.edu.cn, yanxuzhu@bjtu.edu.cn, jtsang@bjtu.edu.cn
}

\usepackage{bibentry}
\begin{document}

\maketitle

\begin{abstract}
The advanced reasoning capabilities of Large Reasoning Models enable them to thoroughly understand and apply safety policies through deliberate thought processes, thereby improving the models' safety. 
Beyond safety, these models must also be able to reflect the diverse range of human values across various cultures. This paper presents the \textbf{C}ultural \textbf{N}orm-based \textbf{C}ultural \textbf{A}lignment~(\textit{\textbf{CNCA}}) framework, which enables models to leverage their powerful reasoning ability to align with cultural norms.
Specifically, we propose three methods to automatically mine cultural norms from limited survey data and explore ways to effectively utilize these norms for improving cultural alignment. Two alignment paradigms are examined: an in-context alignment method, where cultural norms are explicitly integrated into the user context, and a fine-tuning-based method, which internalizes norms through enhanced Chain-of-Thought training data. Comprehensive experiments demonstrate the effectiveness of these methods, highlighting that models with stronger reasoning capabilities benefit more from cultural norm mining and utilization. Our findings emphasize the potential for reasoning models to better reflect diverse human values through culturally informed alignment strategies.
\end{abstract}


\section{Introduction}
Large reasoning models, such as ChatGPT-o1~\cite{jaech2024openai} and DeepSeek-R1~\cite{guo2025deepseek}, exhibit significant advancements in performing complex reasoning tasks, such as mathematical calculations and code generation, due to targeted enhancements in their Chain-of-Thought (CoT) capabilities.
These robust reasoning ability also reshape model alignment; for example, \citet{guan2024deliberative} observe that integrating reasoning with safety policies effectively improves safety alignment.
Beyond safety alignment, given the multicultural nature of our society, it is essential for large models to accommodate and reflect diverse human values and preferences across various cultures~\cite{scherrer2024evaluating,alkhamissi-etal-2024-investigating, wang-cdeval}. 
Consequently, an important research question emerges: 
how to mine and apply culturally specific guidelines—referred to as \textit{Cultural Norms}~\footnote{The definition of cultural norms refers to shared beliefs, or values and the human behaviors that support these values within a given society, such as the standards of conduct that are met with social approval or disapproval. 
% \url{https://study.com/academy/lesson/cultural-norms-definition-values-quiz.html}
}, analogous to safety policies—based on reasoning models, in order to better align with the cultural contexts of specific countries.\par
\begin{figure}[t]
    \centering
    \includegraphics[width=0.9\linewidth]{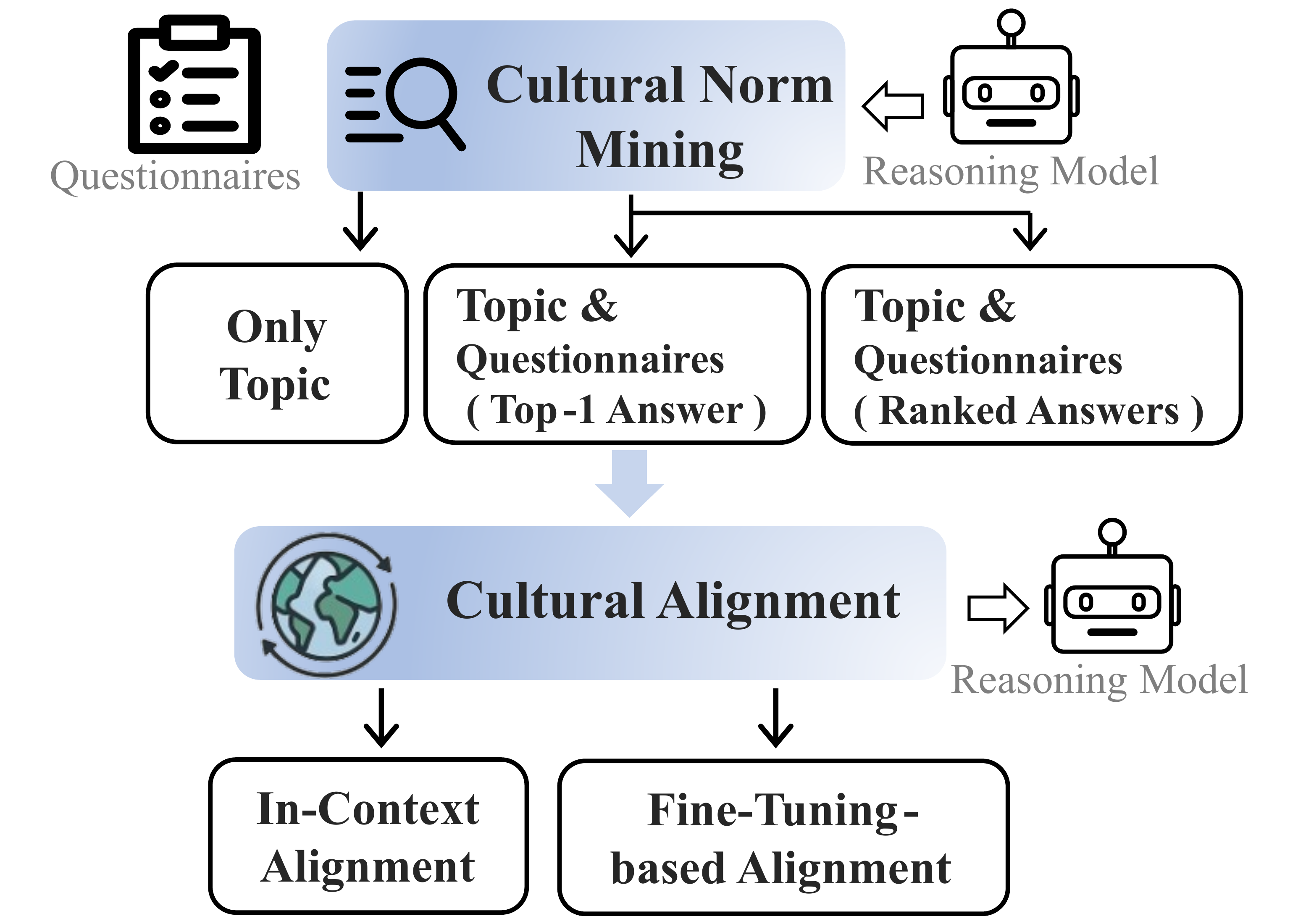}
    \caption{The framework of our proposed \textit{CNCA}.}
    \label{fig:framework}
\end{figure}
\begin{figure}[t]
    \centering
    \includegraphics[width=\linewidth]{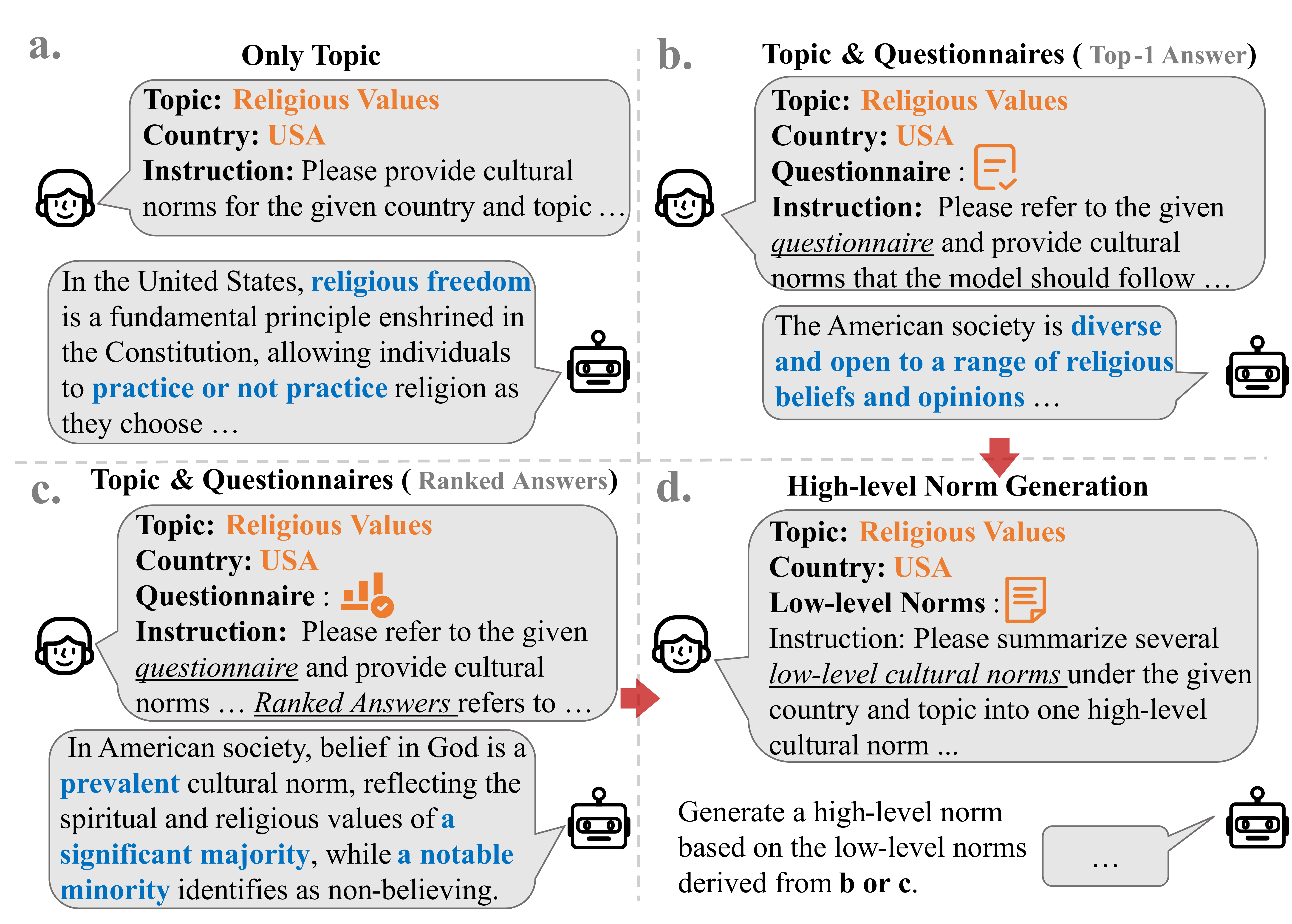} 
    % \vspace{-0.4cm}
    \caption{
Three methods for cultural norm mining:
\textbf{Only Topic~(T)}: Extract norms from the model using only topic information (a);
\textbf{Topic \& Questionnaires Top-1 Answer)~(TQ~(TA))}: Extract norms using both topic information and questionnaire data by selecting top-1 answers~(b$\to$d);
\textbf{Topic \& Questionnaires (Ranked Answers)~(TQ~(RA))}: Similar to TQ~(TA), but based on ranked answers from the questionnaire data~(c$\to$d).
Note that questionnaires represent aggregated survey results from different countries.
Methods TQ~(TA) and TQ~(RA) mine low-level cultural norms, which are then abstracted into higher-level norms.}
    \label{fig:norm-mining}
    % \vspace{-0.4cm}
\end{figure}
In this work, we investigate the \textbf{C}ultural \textbf{N}orm-based \textbf{C}ultural \textbf{A}lignment~(\textit{\textbf{CNCA}}) for reasoning models.
As shown in Figure~\ref{fig:framework}, our study involves two key steps:
% (1) Automatically mining cultural norms from a limited number of cultural survey questionnaires, ensuring practicality in data-scarce scenarios.
(1) Automatically mining cultural norms. 
This step is foundational because ready-made cultural norms are often not readily available and need to be uncovered based on cultural questionnaires.
In Figure~\ref{fig:norm-mining}, we present three consultancy-based cultural norm mining methods. The first method relies solely on topics, whereas the other two methods require a small amount of survey data in addition to topics.
The difference between the two lies in whether the provided answer information is top-1 or ranked, which we will detail in Section \textit{Cultural Norm Mining}.
(2) Exploring methods of utilizing mined cultural norms for cultural alignment. The first method directly incorporates cultural norms as context into user requests, serving as an in-context alignment~\cite{huang-etal-2024-far}. 
The second alignment method follows the fine-tuning paradigm and seeks to internalize cultural norms into the model via norm-enhanced CoT data.
We elaborate on these two methods in Section \textit{Cultural Alignment Methods}. 

Based on the \textit{CNCA} framework, we explore three research questions.  
First, we evaluate the effectiveness of three cultural mining methods within the in-context alignment paradigm. For example, in experiments conducted using DeepSeek-R1-Distill-Qwen-14B \cite{guo2025deepseek},  the norm generated based on \textit{Topic \& Questionnaires (Top-1 Answer)} effectively increases the cultural alignment score by 2.69 compared to the vanilla model (65.55).
Second, we examine the impact of model reasoning capabilities on cultural alignment, generally finding that models with stronger reasoning abilities exhibit better norm mining and utilization.  
Third, we investigate fine-tuning methods for internalizing cultural norms. This allows the model to initiate cultural reasoning on its own without explicitly adding cultural norms in the context. Experimental results show that cultural norm-based fine-tuning outperforms Supervised Fine-Tuning (SFT) and CoT-SFT, and generalizes well to an out-of-distribution, culturally relevant evaluation set, CDEval~\cite{wang-cdeval}.  
In addition to the three research questions above, we discuss the role of cultural norms in non-reasoning models and find that they are still useful, though not as effectively utilized as in reasoning models.
Our contributions are as follows:\par
\begin{itemize}
\item We first position cultural alignment based on cultural norms in the reasoning model, which includes two steps: cultural norm mining and utilizing cultural norms for alignment. The effectiveness of this alignment framework is validated through comprehensive experiments.  
\item  
Building upon the in-context alignment paradigm, we conduct a comparative analysis of three approaches to cultural norm mining and find that combining topic information with top-1 questionnaire answers is most effective. We also show that stronger reasoning ability lead to better cultural alignment.
\item 
For the fine-tuning alignment paradigm, we verify that internalizing cultural norms enhances the model's self-distillation data, thereby improving cultural alignment, which is also validated on out-of-distribution data.
\end{itemize}

\section{Preliminary}

\begin{table}[t]
\centering
% \small
% \renewcommand{\arraystretch}{0.7}
\setlength{\tabcolsep}{0pt} % 修改列间距，使表格更加紧凑
\scalebox{1}{
\begin{tabular}{p{0.99\linewidth}}
\toprule[1.5pt]
\textbf{Topic:} Religious Values \quad \textbf{Country:} USA \\
\midrule
\textbf{\textit{Low-level:}} \textbf{1.} In the United States, religious values, particularly those centered around the importance of God, continue to play a central role in many individuals' lives and cultural identity. \textbf{2.} The cultural norm in the United States reflects a widespread acceptance and integration of religious faith into personal and communal life, alongside recognition of diverse spiritual practices and values, coexisting within a pluralistic society. \textbf{3.} The majority of Americans believe in life after death, reflecting a cultural norm that emphasizes supernatural and religious beliefs. \textbf{4.} In the United States, a significant portion of the population, influenced by predominant Christian values, believes in Hell as a place of damnation for the wicked. \textbf{5.} Belief in Heaven is a significant and widely held cultural norm in American society, reflecting a prevalent religious perspective. \\
\textbf{\textit{High-level:}} Americans hold and express core religious beliefs, particularly centered around concepts of God, life after death, Heaven, and Hell, which significantly influence their understanding of the world and personal values.  \\
\bottomrule[1.5pt]
\end{tabular}}
\caption{An example of cultural norms derived from the \textit{Topic \& Questionnaires ( Top-1 Answer)} method.}
\label{tab: norm case 0}
\end{table}

\subsection{World Values Survey}
The World Values Survey (WVS)~\cite{WVS} is a public opinion survey that collects people's perspectives on 13 cultural topics across different countries. 
The resultant dataset is widely used in cultural studies involving large language models~\cite{alkhamissi-etal-2024-investigating,li2024culturellm,li2024culturepark,xu2024self}. 
The dataset we utilize originates from \cite{xu2024self}, which contains 261 samples. 
For example, under the topic of \textit{Social Values, Attitudes \& Stereotypes}, one of the questionnaire items, \textit{Q1}, asks: \textit{``How important is family in your life?''} The available response options are: \textit{``Very important, Rather important, Not very important, Not at all important''}. 
For instance, in the United States, a significant majority of respondents (approximately 89\%) selected \textit{``Very important''}, while only about 0.3\% chose \textit{``Not at all important''}.
In our study, we adopt the majority response as the ground truth, following the approach commonly used in prior research~\cite{xu2024self}.
Specifically, in our study, for each cultural topic, we select 5 samples to explore cultural norms or use them as a training set.
This results in 65 training samples per country, with the remaining samples reserved for testing.
In our work, we focus on 18 countries, including the USA, Canada, and China.
% (see Appendix for details)
% Table~\ref{tab: wvs dataset} provides examples of cultural topic along with sample questions.
\subsection{Cultural Norm Mining}
\label{sec: cultural_norm_mining}
The analysis of social survey questionnaires to derive meaningful insights constitutes a pivotal methodology in human sociological research. 
% Given that reasoning models possess extraordinary data comprehension capabilities, it is a natural idea to directly utilize reasoning models to automatically uncover cultural norms.
% Given the remarkable data-processing prowess of language models, the integration of AI to augment human analytical capabilities has emerged as a prominent trend.
Considering the remarkable data comprehension ability of reasoning models, this study explores the automatic mining of cultural norms from cultural survey questionnaires using the models themselves.
Our approach involves a specified topic and the results of $m$ cultural surveys (with $m$ set to 5 in our experiments) under that topic. 
We investigate three distinct methods for mining cultural norms, each leveraging different levels of input information, as illustrated in Figure \ref{fig:norm-mining}.
The first method, illustrated in Figure \ref{fig:norm-mining} (a), leverages only the topic information to prompt the model to generate cultural norms for specified countries related to that topic. 
The second method, as shown in Figure \ref{fig:norm-mining} (b) and (d), extends the first approach by incorporating limited survey data along with top-1 answers. In this method, the model first generates low-level norms at the questionnaire level using the provided information (b), and then aggregates these norms to infer higher-level norms (d). 
The third method, as shown in Figure \ref{fig:norm-mining} (c) and (d), is similar to the second method, with the key distinction being the inclusion of ranked answers. Instead of relying solely on the top choice, this method utilizes all options, arranged in descending order according to the questionnaire statistics.
We present an example of a cultural norm mined using the \textit{Topic \& Questionnaires (Top-1 Answer)} method in Table~\ref{tab: norm case 0}.
% For examples from the other two methods, please refer to Table~\ref{tab: norm case} in Appendix. 
% For examples from the other two methods, please refer to the Appendix. 

\begin{figure}[t]
    \centering
    \includegraphics[width=\linewidth]{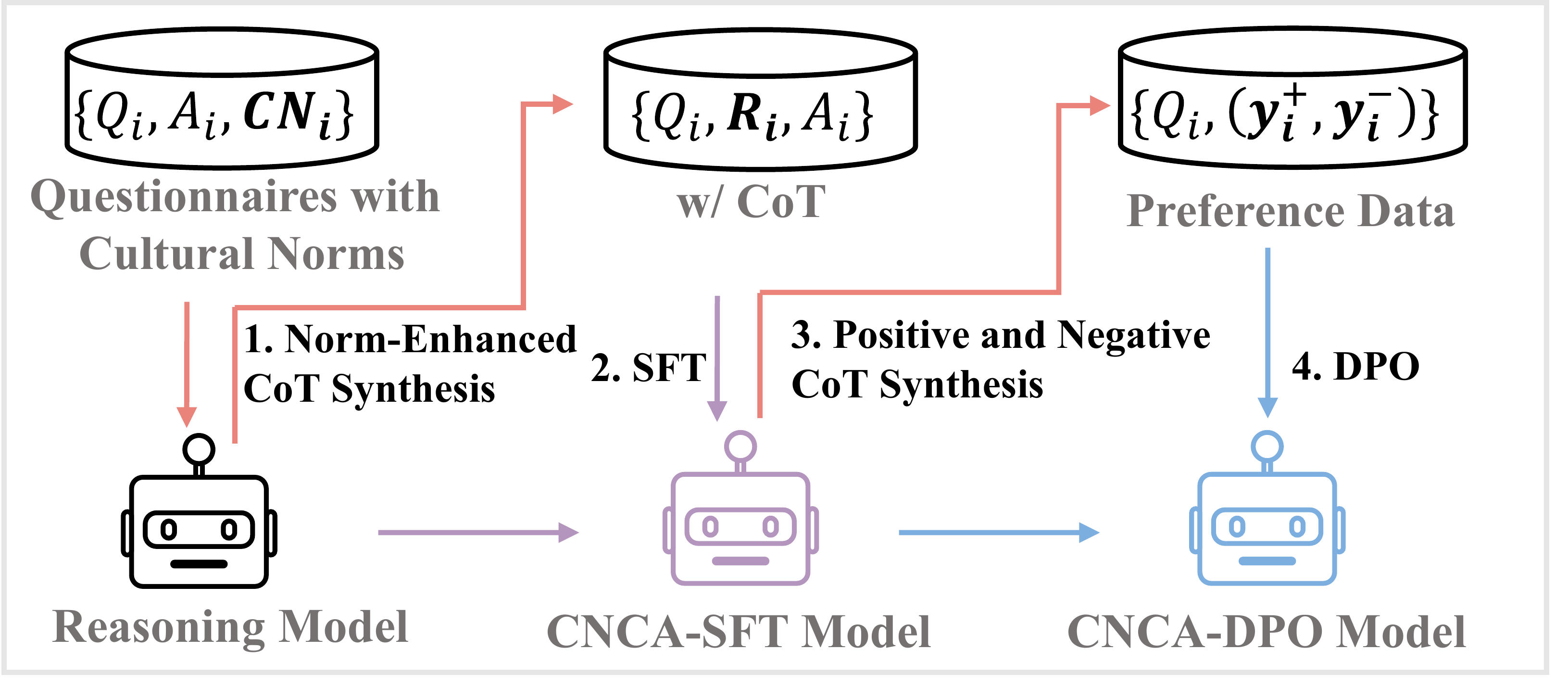}
    % \vspace{-0.5cm}
    \caption{Self-distillation data synthesis and fine-tuning framework based on cultural norms. $CN_{i}$ represents low-level norms, $R_{i}$ denotes reasoning, and $y^+$ and $y^-$ represent correct and incorrect responses with CoT, respectively.}
    \label{fig:finetune-pipeline}
    % \vspace{-0.5cm}
\end{figure}
\begin{table}[t] 
\centering
% \small
% \setlength{\tabcolsep}{0pt} % 修改列间距，使表格更加紧凑
\renewcommand{\arraystretch}{0.98}
\scalebox{1}{
\begin{tabular}{p{0.98\linewidth}}
\toprule[1.5pt]
\textbf{System:} You are a real person with a/an \{\textit{Country}\} cultural background. Please fill out the World Values Survey and answer the questions honestly according to your own value system. \\
\midrule
\textbf{User Instruction~(Standard):}  Given a \#Question and \#Options, choose the option that best aligns with your own value system to answer the question.\\ 
\#Question: \{\} \#Options: \{\}\\
Please return the number of the selected option only.\\
\midrule
\textbf{User Instruction~(In-Context Learning):}  Given a \#Question and \#Options, choose the option that best aligns with your own value system to answer the question. You can refer to the given cases.\\
\textbf{\#Cases: \{\}}\\ 
\#Question: \{\} \#Options: \{\}\\
Please return the number of the selected option only.\\
\midrule
\textbf{User Instruction~(Cultural Norms):}  Given a \#Question and \#Options, choose the option that best aligns with your own value system to answer the question. You can refer to the given cultural norms.\\
\textbf{\#Cultural Norms:} \\
    \quad low-level: \{\}\\
    \quad high-level: \{\}\\
\#Question: \{\} \#Options: \{\}\\
Please return the number of the selected option only.\\
\bottomrule[1.5pt]
\end{tabular}}
\caption{Prompt for testing model cultural alignment. Note that all fine-tuned models use the standard user instruction as the prompt.}
% \vspace{-1cm}
\label{tab: test-prompt}
\end{table}

\subsection{Cultural Alignment Methods }
\label{sec: cultural_alignment_methods}
In this work, we explore two types of cultural alignment methods.
The first type is in-context alignment, which directly integrates the extracted cultural norms corresponding to the topic of the test question into the user's request. This approach is similar to In-Context Learning (ICL), but differs in that ICL typically involves inserting specific questionnaire examples into the context. 
% We compared these two methods in the third and fourth rows of Table~\ref{tab: test-prompt}, and the example of cultural norms used can be found in Appendix Table~\ref{tab: norm case}.
% We compared these two methods in the third and fourth rows of Table~\ref{tab: test-prompt}, and the example of cultural norms used can be found in Appendix.\par
We compared these two methods in the third and fourth rows of Table~\ref{tab: test-prompt}.\par
The second method relies on fine-tuning. 
As shown in Figure~\ref{fig:finetune-pipeline}, we first synthesize thought processes (\{$R_i$\}) by self-distilling the vanilla model using questionnaires (\{$Q_{i},A_{i}$\}) and mined question-level cultural norms (\{$CN_{i}$\}).
This approach improves upon standard CoT-SFT by incorporating cultural norms into the thinking process, enhancing its quality.
% The prompts follow the in-context alignment, specifically the fourth row in Table~\ref{tab: test-prompt}. 
For each sample, we generate 10 reasoning chains, each with 10 responses, until the correct answer is found. If unsuccessful, we retain the sample without reasoning. We then perform SFT on the vanilla model using the collected data, omitting cultural norms from the instructions. The SFT loss is defined in Equ~\ref{equ:sft loss}. Next, we use the \textit{CNCA-SFT} model to generate positive and negative samples under similar settings, again excluding norms from instructions. Finally, we conduct Direct Preference Optimization (DPO) using these samples, with the loss shown in Equ~\ref{equ:dpo loss}.

\begin{equation}
\mathcal{L}_{\text{SFT}}(\theta) = 
-\mathbb{E}_{\substack{(Q_i,R_i,A_i) \\ \sim \mathcal{D}}}
\Bigl[ 
\log P_{\theta}(R_i,A_i \mid Q_i) 
\Bigr]
\label{equ:sft loss}
\end{equation}

\begin{equation}
\begin{split}
\mathcal{L}_{\text{DPO}}(\theta) = 
-\mathbb{E}_{\substack{(Q_i,y^+_i,y^-_i) \\ \sim \mathcal{D}}}
& \log \sigma \biggl[ \beta \biggl( 
\log \frac{P_\theta(y^+_i|Q_i)}{P_{\text{ref}}(y^+_i|Q_i)} \\
& \quad - \log \frac{P_{\theta}(y^-_i|Q_i)}{P_\text{ref}(y^-_i|Q_i)}
\biggr) \biggr]
\end{split}
\label{equ:dpo loss}
\end{equation}

\section{Research Questions}
% \noindent\textbf{The Effectiveness of Cultural Norm Mining Methods.} 
\noindent\textbf{Which Cultural Norm Mining Method is the Most Effective?} 
Cultural norm mining is a foundational step in \textit{CNCA}, as the cultural norms identified have a direct impact on the effectiveness of cultural alignment, highlighting the importance of selecting the appropriate methods.
Based on the in-context alignment paradigm, we evaluate three cultural norm mining methods (illustrated in Figure \ref{fig:norm-mining}) across reasoning models of varying scales. 
This training-free approach enables us to intuitively assess the effectiveness of these methods by observing the alignment outcomes.\par
% \noindent\textbf{Reasoning Ability Affects Cultural
% Alignment.} 
\noindent\textbf{How Does Reasoning Ability Affect Cultural Alignment?} 
Within the \textit{CNCA} framework, both the mining and utilization of cultural norms are related to the model's reasoning capabilities. We aim to explore the impact of the model's reasoning ability in these two aspects on alignment based on in-context alignment.
This exploration will guide us in developing more effective methods for the future.
\par
% \noindent\textbf{Fine-tuning with Cultural Norms Enhanced Chain-of-Thought Data.}
\noindent\textbf{Can Cultural Norms Be Internalized in Models?}
While cultural norms can be directly incorporated through in-context alignment, our goal is to internalize these norms into the model through enhanced Chain-of-Thought (CoT) fine-tuning. This approach allows the model to autonomously invoke cultural reflections during application, without the need for explicit cultural norm inputs, thereby minimizing unnecessary token consumption.
\par

\section{Experimental Setup}
\noindent\textbf{Large Reasoning Models.}
In this study, we explore three large reasoning models: DeepSeek-R1-Distill-Qwen-7B, DeepSeek-R1-Distill-Llama-8B, and DeepSeek-R1-Distill-Qwen-14B. These models are derived through supervised fine-tuning on a curated dataset of 800k entries distilled from DeepSeek-R1~\cite{guo2025deepseek}, with Qwen2.5-Math-7B, Llama-3.1-8B, and Qwen2.5-14B serving as the base models~\cite{yang2024qwen2,grattafiori2024llama, yang2024qwen2a}. 
Following the distillation process, the models retain the CoT capabilities of DeepSeek-R1. During inference, their reasoning process is structured and generated using the \texttt{$<$think$>$ $<$/think$>$} tags. Note that for the sake of convenience in description, we will use R1-Qwen-7B, R1-Llama-8B, and R1-Qwen-14B to denote the aforementioned reasoning models respectively.\par
\noindent\textbf{Comparison methods.}\par
\begin{itemize}
    \item In-Context Learning~(ICL): This method involves directly providing the model with questionnaire examples within the user's context during inference. The prompt format is shown in Table~\ref{tab: test-prompt}.
    \item Supervised Fine-Tuning~(SFT): The model is fine-tuned on the training dataset without incorporating any intermediate reasoning processes. 
    \item CoT based SFT~(CoT-SFT): 
    CoT-SFT enhances standard SFT by requiring the model to self-distill CoT from the ground truth for better process supervision. We perform 10 rounds of sampling, generating 10 responses per round. If no valid reasoning leading to the correct answer is found within the predefined rounds, the standard data format is used without reasoning steps.
\end{itemize}

\noindent\textbf{Cultural Alignment Metric.}
The WVS consists of N multiple-choice questions with numerical response options (e.g., 1: Strongly Disagree, 2: Disagree, 3: Neutral, etc.).
For a given country $c$, we use the majority choices of the respondents as the ground truth to construct the cultural answer vector: 
$A_c = [a^c_1, a^c_2, ..., a^c_N]$.
Next, we prompt the model to answer these questions, producing the model's output vector 
$R_c = [r^c_1, r^c_2, ..., r^c_N]$. 
Following \citet{wang-etal-2024-countries} and \citet{xu2024self}, we calculate the cultural alignment score $S(A_c, R_c)$ as follows:
\begin{equation}
   S(A_c, R_c) = \left(1 - \frac{\sqrt{\sum_{i=1}^{N} (a^c_i - r^c_i)^2}}{\text{max\_distance}}\right) \times 100 
\end{equation}
where max\_distance is the maximum possible difference between selected options, ensuring normalization. A higher score indicates greater alignment with country $c$.
Note that the results reported in the paper are the average score obtained from 18 countries. 
% For the results of each individual country, please refer to the Table~\ref{tab: detailed result part0},~\ref{tab: detailed result part1},~\ref{tab: detailed result part2} in Appendix. 
% For the results of each individual country, please refer to the Appendix. 
Additionally, each result presented is the average of three experimental trials.
\par
\noindent\textbf{Technical Details.} 
For inference, we employ the vLLM~\cite{kwon2023efficient} efficient inference framework, setting the temperature to 0.6 and the maximum sequence length to 1024, while keeping all other parameters at their default values. All reported experimental results represent the average over three independent runs.  
For training, we utilize the LoRA~\cite{Hu2021LoRALA} efficient parameter fine-tuning method. All experiments undergo a single epoch of training, with the learning rate set to 5e-5, a warm-up phase covering 10\% of the total training steps, and a cosine learning rate scheduler.
The training of the 14B model is conducted on a single 80G A800 device, while all other experiments are performed using two 24G RTX 3090 GPUs. 
\par

\section{Results}
% 文化norm三种挖掘方式
% 同一种挖掘方式下：
    % 模型生成norm的能力
    % 模型利用norm的能力
%消融：不同norm个数使用情况的分析
%提升模型利用norm的能力，sft dpo
\begin{table}[t]
\centering
% \small
% \setlength{\tabcolsep}{3pt} % 调整列间距
\renewcommand{\arraystretch}{0.4}
\begin{tabular}{l|l|l}
\toprule
\textbf{Model} & \textbf{Method} & \textbf{Score~($\uparrow$)} \\
\midrule
\multirow{7}{*}{\textbf{\makecell{R1-Qwen-7B}}}
    & \textit{Vanilla} & 61.22 \\
    & \textit{ICL} & 60.38 \\
    & \textit{SFT} & 61.87 \\
    & \textit{CoT-SFT} & \textbf{62.48} \\
    \cmidrule(lr){2-3}
    &  CNCA-T &  61.19~(-0.03) \\
    &  CNCA-TQ~(TA) &  61.66~(+0.44) \\
    &  CNCA-TQ~(RA) &  61.11~(-0.11) \\
\midrule
\multirow{7}{*}{\textbf{\makecell{R1-Llama-8B}}} 
    & \textit{Vanilla} & 64.13 \\
    & \textit{ICL} & 62.05 \\
    & \textit{SFT} & 64.91 \\
    & \textit{CoT-SFT} & 61.28 \\
   \cmidrule(lr){2-3}
    &  CNCA-T &  64.77~(+0.64) \\
    &  CNCA-TQ~(TA) &  \textbf{65.80}~(+1.67) \\
    &  CNCA-TQ~(RA) &  64.12~(-0.01) \\
\midrule
\multirow{7}{*}{\textbf{\makecell{R1-Qwen-14B}}} 
    & \textit{Vanilla} & 65.55 \\
    & \textit{ICL} & 67.21 \\
    & \textit{SFT} & 65.99 \\
    & \textit{CoT-SFT} & 66.20 \\
    \cmidrule(lr){2-3}
    &  CNCA-T &  66.40~(+0.85) \\
    &  CNCA-TQ~(TA) &  \textbf{68.24}~(+2.69) \\
    &  CNCA-TQ~(RA) &  65.94~(+0.39) \\
\bottomrule
\end{tabular}
\caption{Experimental results of in-context cultural alignment based on cultural norms.  
The best results within each group are highlighted in bold.}
% \vspace{-0.5cm}
\label{tab:model_performance}
\end{table}

\begin{figure}[t]
    \centering
    \includegraphics[width=\linewidth]{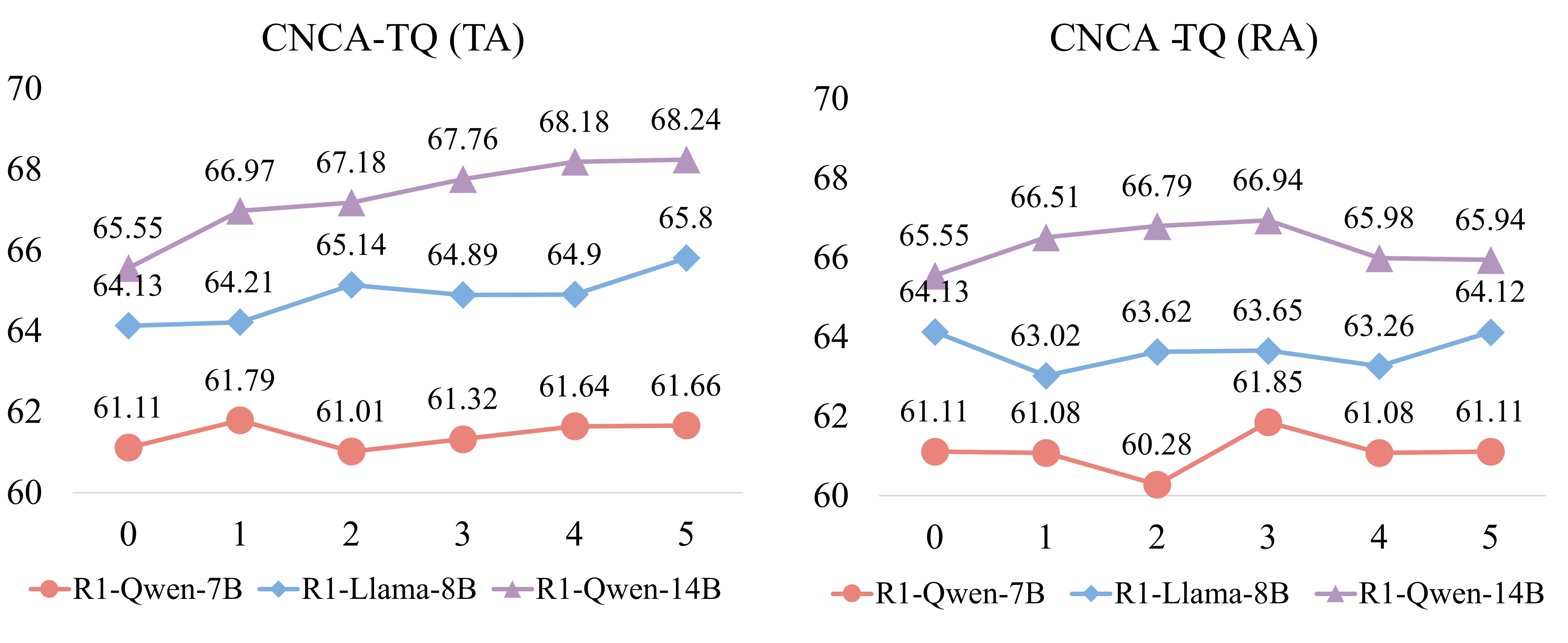}
    % \vspace{-0.5cm}
    \caption{The impact of the number of cultural norms (x-axis) on alignment performance (y-axis). The left and right plots correspond to the experimental results of CNCA-TQ~(TA) and CNCA-TQ~(RA), respectively.}
    \label{fig:ablation}
    % \vspace{-0.5cm}
\end{figure}
\subsection{The Effectiveness of Cultural Norms Mining Methods} 
The experimental results is shown in Table~\ref{tab:model_performance}.  
For the sake of convenience in description, we respectively
use \textit{CNCA-T}, \textit{CNCA-TQ~(TA)}, \textit{CNCA-TQ~(RA)} to denote cultural alignment based on three cultural norm mining methods.
The methods exhibit varying degrees of effectiveness, with notable differences among them. The performance of each method scales positively with the increase in model size. \par
Specifically, \textit{CNCA-TQ~(TA)} emerges as the most effective, demonstrating significant enhancements over the vanilla model across all three evaluated models. Particularly on R1-Llama-8B and R1-Qwen-14B, \textit{CNCA-TQ~(TA)} achieves optimal results, with improvements of 1.67 and 2.69 points, respectively. 
The corresponding p-values from the t-tests are 0.021 and 0.003, confirming that the improvements are statistically significant.
% (see Appendix for detailed results)
However, its performance on R1-Qwen-7B is inferior to that of the two fine-tuning-based baselines, a discrepancy attributed to the foundational model's reasoning capabilities, which will be discussed in detail in the Section \textit{Reasoning Ability Affects Cultural Alignment}.
\textit{CNCA-T} ranks second in terms of performance. 
Despite its reliance on norms solely derived from topic information, this method yields commendable results. 
For instance, in the R1-Llama-8B model, it is only marginally outperformed by the ICL baseline, which benefits from the inclusion of a limited number of examples. 
We posit that \textit{CNCA-T} offers norms that are more generalized yet less specifically targeted.\par
Conversely, \textit{CNCA-TQ~(RA)} is the least effective, showing only a slight improvement on R1-Qwen-14B.
In fact, although the ranked answers input during the cultural norm mining process provide more information, the most important answer in cultural alignment is usually the top-1 answer. Extra information may introduce noise, potentially leading the model astray. 
% We provide examples of norms in Figure~\ref{fig:norm-mining} and in the Appendix.\par
We provide examples of norms in Figure~\ref{fig:norm-mining}.\par
Futhermore, we conduct experiments on the impact of the number of norms on cultural alignment. The experimental results are shown in Figure~\ref{fig:ablation}. We find that for  \textit{CNCA-TQ~(TA)}, the performance of the three models steadily improves as the number of norms increases. However, for \textit{CNCA-TQ~(RA)}, the trends of the three models differ. The R1-Qwen-7B and R1-Qwen-14B models achieve their best performance when the number of norms is 3, surpassing the results reported in the Table~\ref{tab:model_performance}, while the R1-Llama-8B model performs best at the extremes. 
This also indicates that the additional (redundant) information introduced based on \textit{CNCA-TQ~(RA)} can cause the model to exhibit some confusion between useful information and noise descriptions, thereby leading to instability in the model's performance.

\begin{figure}[t]
    \centering
    \includegraphics[width=\linewidth]{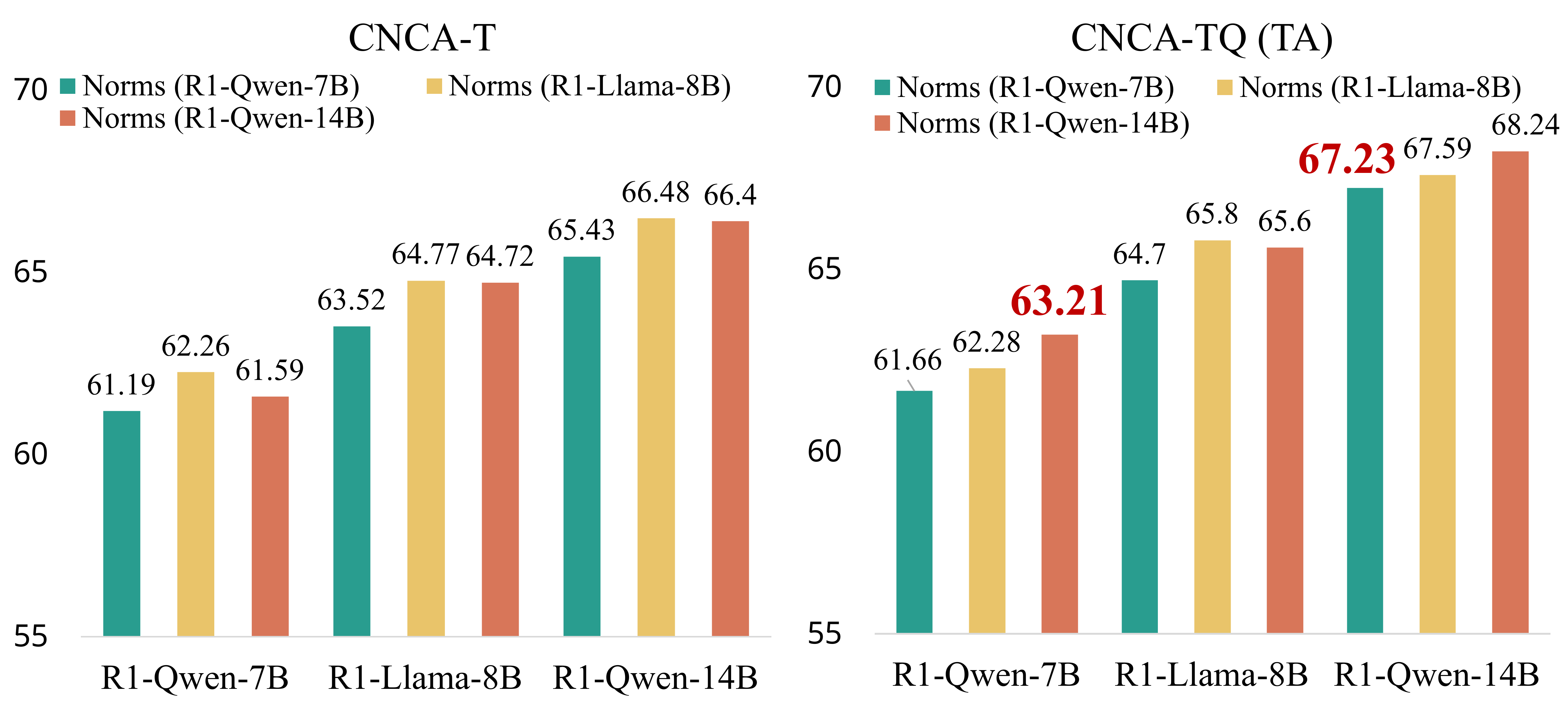}
    % \vspace{-0.4cm}
    \caption{Results of cultural alignment evaluations based on norms generated by various models. The x-axis represents the inference models, the y-axis indicates the alignment scores, and the colors distinguish norms originating from different models.}
    % \vspace{-0.2cm}
    \label{fig:cross-test}
\end{figure}

\subsection{Reasoning Ability Affects Cultural Alignment}
\label{sec: factor}
From Table~\ref{tab:model_performance}, we observe that as the model scale increases, the alignment effectiveness of \textit{CNCA} improves significantly, which we attribute to the model's reasoning capabilities.
This is reflected in two aspects within \textit{CNCA}: firstly, the model's ability to infer cultural norms based on questionnaires, and secondly, the model's ability to align using these inferred cultural norms. Given this, under the in-context alignment setting, we test each model using norms derived from different models, with the results shown in Figure~\ref{fig:cross-test}. Note that our experiments are based on \textit{CNCA-T} and \textit{CNCA-TQ~(TA)}, as they are the two methods with better performance.

% Firstly, when R1-Qwen-7B uses norms generated by larger models, its alignment performance improves significantly across two cultural mining methods.
Firstly, we fix the evaluation model and observe the results of using different models to generate norms. For example, when R1-Qwen-7B employs norms generated by larger models, the alignment performance of both cultural mining methods improves significantly. Under the \textit{CNCA-T} method, the norms produced by R1-Lama-8B yield the best results, while R1-Qwen-14B performs comparably.  
For the \textit{CNCA-TQ (TA)} method, the norms generated based on R1-Qwen-14B deliver the best performance.
This difference arises because \textit{CNCA-T} relies only on topics and demands less reasoning, while \textit{CNCA-TQ~(TA)} requires analyzing examples, thus needing stronger reasoning capabilities.
These findings support our first hypothesis: models with greater capabilities produce higher-quality cultural norms.

Secondly, in experiments using the \textit{CNCA-TQ~(TA)} method, R1-Qwen-14B achieves a score of 67.23 when using norms generated by the smaller R1-Qwen-7B. Although this is slightly lower than when using its own norms, it still outperforms the best baseline ICL performance (67.21). This demonstrates that models with stronger reasoning abilities can effectively utilize norms of slightly lower quality, showing greater robustness to norm quality.

\subsection{Fine-tuning with Cultural Norms Enhanced Chain-of-Thought Data \label{sec: finetuning}}
\begin{table}[t]
\centering
% \small
\renewcommand{\arraystretch}{0.6}

\begin{tabular}{l|c}
\toprule[1.5pt]
Method & Score~($\uparrow$)  \\
\midrule
Vanilla & 65.55  \\
SFT & 65.99 \\
CoT-SFT & 66.20 \\
\midrule
CNCA-SFT & 66.73~(+1.18)\\
CNCA-DPO & 66.90~(+1.35)\\
\bottomrule[1.5pt]
\end{tabular}
% \vspace{-0.2cm}
\caption{Experimental results of the fine-tuning approach based on cultural norms, using the R1-Qwen-14B backbone model.}
% \vspace{-0.2cm}
\label{tab: finetuning-result}
\end{table}
\begin{table}[t]
\centering
\small
\setlength{\tabcolsep}{3pt} % 修改列间距，使表格更加紧凑

\begin{tabular}{l|c|c|c|c|c|c|c}

\toprule[1.5pt]
Method & LTO & MAS & PDI & IDV & IVR & UAI & Avg.  \\
\midrule
Vanilla & 92.94  &	32.25 &	58.83 &	59.96 &	52.39 &	57.03 &	58.90 \\
SFT & \textbf{93.25}	& 35.01 & \textbf{60.74} & 60.49 & 51.84 & 57.74 & 59.84\\
CoT-SFT & 93.04 & 36.28 & 59.71	& 60.59 & 54.45 & 58.20 & 60.38\\
\midrule
CNCA-SFT & 92.58 & 34.57 & 60.06 & 60.75 & 53.74 & \textbf{58.58} & 60.05\\
CNCA-DPO & 91.08 & \textbf{37.56 } & 60.69 & \textbf{62.08}	& \textbf{55.08} & 56.65 & \textbf{60.52}\\
\bottomrule[1.5pt]
\end{tabular}
% \vspace{-0.2cm}
\caption{Experimental results of the fine-tuning method based on cultural norms on the CDEval dataset.}
\label{tab: ood-result}
\end{table} 
\begin{table}[t] 
\centering
\small
\setlength{\tabcolsep}{0pt} % 修改列间距，使表格更加紧凑
\renewcommand{\arraystretch}{1}
\scalebox{1}{
\begin{tabular}{p{0.98\linewidth}}
\toprule[1.5pt]
\textbf{Question:} Would you not like to have drug addicts as neighbors? \\
\textbf{Options:} 1. Yes, i would not like to \quad 2. No, i would like to\\ 
\textbf{Country:} USA \quad \textbf{Ground Truth:} 1 \\
\midrule
\textbf{Vanilla:} $<$think$>$ \textbf{...} First, I should consider my personal feelings and beliefs about this. 
\textbf{...} Another angle is the principle of empathy and compassion. \textbf{...} 
\textbf{\textit{Therefore, I might choose No, indicating that I wouldn't mind having them as neighbors, especially if it's in a supportive environment.}} $<$/think$>$ 2 \\
\midrule
\textbf{SFT:} $<$think$>$ First, I should consider my personal feelings about this. \textbf{...} I should reflect on my own values. I value compassion and helping people, so I might be inclined towards option 2.
\textbf{...}
Considering my cultural background, which emphasizes individualism and helping the less fortunate, I might lean towards wanting to help rather than exclude.  \textbf{...}
\textbf{\textit{I should go with my initial gut feeling, which is more towards compassion and helping, so option 2.}} $<$/think$>$ 2 \\
\midrule
\textbf{CoT-SFT:} $<$think$>$ \textbf{...} First, I should consider my personal feelings about having drug addicts as neighbors. I know that drug addiction is a complex issue and can lead to various problems like health risks,  \textbf{...}
\textbf{\textit{I would say ``No, I would like to,'' because I believe in giving people another chance and supporting their recovery.}}  $<$/think$>$ 2 \\
\midrule
\textbf{CNCA-DPO:} $<$think$>$ \textbf{...} \textbf{\textit{First, I should consider my personal feelings and values. Growing up in the U.S., I've been influenced by cultural norms that emphasize health, safety, and community well-being.}} Drug addiction is often seen as a serious issue, \textbf{...},  \textbf{\textit{I would lean towards not wanting drug addicts as neighbors because of the potential negative impacts on the community and personal safety}}, even though I support helping them through treatment and support systems.$<$/think$>$ 1 \\

\bottomrule[1.5pt]
\end{tabular}}
\caption{Examples of outputs from the Vanilla, SFT, COT-SFT, and CNCA-DPO models.}
% \vspace{-0.5cm}
\label{tab: case study}
\end{table}

% Although the in-context alignment method achieves commendable results, it requires the inclusion of cultural norms in the user's input for each sample. 
In this subsection, we explore an internalized cultural norm fine-tuning approach, aiming to eliminate the need for inputting corresponding cultural norms during model testing. 
Our method, 
introduced in Section \textit{Cultural Alignment Methods}, 
% introduced in \textbf{Section: Cultural Alignment Methods}, 
specifically employs \textit{Topic \& Questionnaires (Top-1 Answer)} to mine norms and utilizes low-level norms due to their close correspondence with the samples themselves. 
As presented in Table~\ref{tab: finetuning-result}, the SFT and DPO variants of the \textit{CNCA} method achieve performance gains of 1.18 and 1.35, respectively, compared to the vanilla model. Furthermore, they outperform the strongest fine-tuning-based baseline, CoT-SFT, by margins of 0.53 and 0.70, respectively. These results validate that incorporating cultural norms facilitates the model’s reasoning process via self-distillation, thereby enhancing overall alignment performance.
This validation shows that cultural norms assist the model in enhancing the quality of its reasoning through self-distillation, thereby improving alignment effectiveness.
In Table~\ref{tab: case study}, we present sample outputs from the model fine-tuned based on cultural norms. It can be observed that 
\textit{CNCA-DPO} method triggers reflections on the specified country's cultural norms during the thought process, which aids the model in decision-making. This also demonstrates that the method can internalize cultural norms within the model.\par 
To verify the generalization of our method,
we further evaluate it on the out-of-distribution culturally relevant dataset CDEval~\cite{wang-cdeval}, which covers six cultural dimensions: Power Distance Index (PDI), Individualism (IDV), Uncertainty Avoidance (UAI), Masculinity (MAS), Long-term Orientation (LTO), and Indulgence vs. Restraint (IVR). 
Each sample provides two options, e.g., in PDI, one reflects high power distance, the other low. Cultural alignment is measured by comparing model predictions with human tendencies using: 
% \begin{align}
%     M^{(i)}_{\mathit{CD}}(c, d) &= \mathcal{\mathbb{1}}[a^{(i)}_{c,d} = g_{c,d}],  \\
%     S_{\mathit{CD}}(d) &= \frac{1}{|C|} 
%     \sum_{c \in C} 
%     \bigg( \frac{1}{N_{cd}} 
%     \sum_{i=1}^{N_{cd}} M^{(i)}_{\mathit{CD}}(c, d) \bigg) , 
% \end{align}
\begin{align}
    M^{(i)}_{\mathit{CD}}(c, d) &= \mathbf{1}[a^{(i)}_{c,d} = g_{c,d}],  \\
    S_{\mathit{CD}}(d) &= \frac{1}{|C|} 
    \sum_{c \in C} 
    \bigg( \frac{1}{N_{cd}} 
    \sum_{i=1}^{N_{cd}} M^{(i)}_{\mathit{CD}}(c, d) \bigg) , 
\end{align}
where $ g_{c,d}$  denotes the cultural tendency of country $c$ in dimension $d$, and $ a_{c,d} $ represents the model's cultural tendency in dimension $ d $ when adopting the identity of country $c$. If these two values align, it is recorded as 1.
Table~\ref{tab: ood-result} reports average scores for China, Germany, the United States, and Russia. 
% Our \textit{CNCA-DPO} achieves the best performance, while \textit{CNCA-SFT} lags behind the self-distillation-based CoT-SFT, highlighting the superior generalization of \textit{CNCA-DPO}.
Our \textit{CNCA-DPO} achieves the best performance, while \textit{CNCA-SFT} also performs well, ranking just behind CoT-SFT, demonstrating the generalizability of our proposed approach.

\begin{table}[t]
\centering
% \small
% \renewcommand{\arraystretch}{0.7}
% \setlength{\tabcolsep}{10pt} % 修改列间距，使表格更加紧凑
\scalebox{1}{
\begin{tabular}{l|l}
\toprule[1.5pt]
\textbf{Model/Method} & \textbf{Score~($\uparrow$)}  \\
\midrule
\textbf{R1-Llama-8B } & 64.13  \\
+ CNCA-TQ~(TA) & \textbf{65.60}~(+1.47)\\
\textbf{Meta-Llama-3.1-8B-Instruct} & 62.96 \\
+ CNCA-TQ~(TA) & 64.10~(+1.14)\\
\midrule
\textbf{R1-Qwen-14B } & 65.55  \\
+ CNCA-TQ~(TA) & 68.24~(+2.69)\\
\textbf{Qwen2.5-14B-Instruct} & 66.56 \\
+ CNCA-TQ~(TA) & \textbf{68.86}~(+2.30)\\
\bottomrule[1.5pt]
\end{tabular}}
\caption{Comparison of cultural alignment results of reasoning and non-reasoning models}
\label{tab:sys1}
% \vspace{-0.5cm}
\end{table}

\section{Discussion \label{sec: sys12}}
Non-reasoning models significantly influence the development of language models. In contrast to System-2~(reasoning models), they rely more on ``intuition'' for decision-making~\cite{li2025system}. This subsection compares the application of cultural norms between the two type of models.
Since the above experiments verify that the norms mined using the \textit{Topic \& questionnaire (Top-1 answer)} based on the R1-Qwen-14B model yield the best results, we conduct comparative experiments based on this approach.
For models, we select Meta-Llama-3.1-8B-Instruct and Qwen2.5-14B-Instruct as representatives of non-reasoning models\footnote{We does not conduct experiments on the R1-Qwen-7B model because the corresponding math model's instruction-following capability is insufficient, making it impossible to extract valid responses from the model's outputs, thus lacking the conditions for experimentation.}.
The experimental results are detailed in Table \ref{tab:sys1}. For the Llama-8B model, the reasoning model consistently outperforms the non-reasoning model irrespective of whether cultural norms are incorporated. Conversely, in comparisons involving the Qwen-14B models, we observe that when cultural norms are not applied, the R1-Qwen-14B underperforms relative to the non-reasoning counterpart. This discrepancy may stem from a ``forgetting'' phenomenon associated with post-training procedures. Nonetheless, significant improvements emerge for the reasoning-based model once cultural norms are introduced.
In summary, reasoning models demonstrate a superior capability in leveraging cultural norms.
% However, this advantage diminishes as model size increases, primarily due to the enhanced intrinsic comprehension ability of larger non-reasoning models.

\section{Related Work}
\subsection{Cultural Alignment}
Studies on exploring the cultural alignment of large language models can be broadly categorized into two main categories. 
The first focuses on the evaluation and analysis of models' culture. For example, comparing LLM outputs with results from human social surveys, previous works \cite{cao-etal-2023-assessing,alkhamissi-etal-2024-investigating,wang-etal-2024-countries,naous2023having,chiu2024culturalbench,masoud2023cultural,wang2023not,Rao2024NormAdAF} reveal that these models exhibit limited cultural adaptability. \par
% Notably, most large language models demonstrate stronger alignment with the cultural norms of English-speaking countries.
The second category emphasizes the automatic construction and expansion of culture-related data based on the language model's own capabilities to finetune models. For instance, \citet{li2024culturellm} introduces a semantic data augmentation approach designed to enhance human questionnaire datasets. Additionally, \citet{li2024culturepark} and \citet{xu2024self} propose generating supervised fine-tuning data through self-instruct techniques and multi-agent dialogue methods, respectively.

\subsection{Principle Learning}
Principle learning represents a body of work that utilizes guidelines derived from human input or model-driven exploration to better accomplish specific tasks. Notable examples include Constitutional AI
~\cite{bai2022constitutional} and SELF-ALIGN~\cite{sun2023principle}, which rely on human-defined rules to enable scalable supervision and model alignment. Another category of work involves the automatic discovery of principles by models to complete designated tasks. For instance, \citet{zhang2024context} and \citet{sun2024retrieved} enhance  models' mathematical performance by incorporating learned principles into in-context learning.
\subsection{Large Reasoning Models}
To improve large models' performance in complex reasoning tasks, significant research has focused on CoT techniques. \citet{wei2022chain} introduced CoT prompting to boost LLMs' reasoning capabilities, leading to a proliferation of new prompting methods~\cite{zhou2022least}. Concurrently, studies have explored enhancing reasoning without explicit prompts using process reward models, advanced search algorithms, and reinforcement learning~\cite{lightman2023let, yao2023tree, kumar2024training, jaech2024openai}. 
A notable advancement is OpenAI's o1 and o3 series~\cite{jaech2024openai}, which achieves extended reasoning through scaled CoT outputs.
Recent large reasoning models, including DeepSeek series~\cite{guo2025deepseek,liu2024deepseek}, Gemini-2.0 \cite{gemini_flash_thinking}, QWQ-32B-Preview~\cite{qwq-32b-preview}, , and Kimi-v1.5~\cite{team2025kimi}, leverage advanced reasoning architectures to enhance their cognitive capabilities.

\section{Conclusion}
This paper investigates the Cultural Norm-based Cultural Alignment (\textbf{CNCA}) framework to enhance cultural alignment in large reasoning models. Experiments demonstrate the effectiveness of mining cultural norms from limited survey data and applying them through in-context and fine-tuning alignment methods. 
Models with strong reasoning capabilities particularly benefit from the integration of cultural norms. Additionally, the cultural norms proposed in this work can provide inspiration for future scalable supervision. Similar self-mined guidelines can serve as an effective signal to improve the quality of model supervision, thereby helping models continuously enhance their alignment performance.

\appendix

\section{Acknowledgments}
We thank the anonymous reviewers for their valuable comments.
This work is supported by the Fundamental Research Funds for the Central Universities (No. 2025JBZX057) and the National Natural Science Foundation of China (No. 62172094, 62576030).

% \bigskip
% \noindent Thank you for reading these instructions carefully. We look forward to receiving your electronic files!

\bibliography{aaai2026}

\end{document}